\documentclass[preprint,5p,times,twocolumn]{elsarticle}
\usepackage{amssymb}
\usepackage{subfigure}
\usepackage{array}
\usepackage[table,xcdraw]{xcolor}
\usepackage[normalem]{ulem}
\useunder{\uline}{\ul}{}
\usepackage{float}
\usepackage[ruled,vlined]{algorithm2e}
\usepackage{setspace}
\usepackage{enumitem}
\usepackage{hyperref}
\usepackage{amssymb}
\usepackage{caption}
\usepackage{tabularx}
\usepackage{pifont}

\journal{}
\begin{document}
\captionsetup[figure]{labelfont={bf},labelformat={default},labelsep=period,name={Fig.}}

\begin{frontmatter}
\title{MSVQ: Self-Supervised Learning with Multiple Sample Views and Queues}

\author{Chen Peng}
\author{Xianzhong Long \corref{cor1}}
\ead{lxz@njupt.edu.cn}
\author{Yun Li} 
\address{School of Computer Science, Nanjing University of Posts and Telecommunications, Nanjing, 210023, China}
\cortext[cor1]{Corresponding author}

\begin{abstract}
Self-supervised methods based on contrastive learning have achieved great success in unsupervised visual representation learning. However, most methods under this framework suffer from the problem of false negative samples. Inspired by the mean shift for self-supervised learning, we propose a new simple framework, namely Multiple Sample Views and Queues (MSVQ). We jointly construct three soft labels on-the-fly by utilizing two complementary and symmetric approaches: multiple augmented positive views and two momentum encoders that generate various semantic features for negative samples. Two teacher networks perform similarity relationship calculations with negative samples and then transfer this knowledge to the student network. Let the student network mimic the similarity relationships between the samples, thus giving the student network a more flexible ability to identify false negative samples in the dataset. The classification results on four benchmark image datasets demonstrate the high effectiveness and efficiency of our approach compared to some classical methods. Source code and pretrained models are available \href{https://github.com/pc-cp/MSVQ}{here}.
\end{abstract}
\end{frontmatter}
\section{Introduction}\label{1}
Self-supervised learning (SSL) has received sufficient attention and rapid progress in the computer vision community. This is due to its ability to learn rich semantic features using unlabeled data \cite{40tian2020contrastive, 27hjelm2018learning, 32lai2019contrastive, 6chen2020simple, 9chen2021exploring, 47wu2018unsupervised, 23grill2020bootstrap, 24he2020momentum}. 
Early self-supervised learning methods were typically based on geometric transformations or heuristics to design the corresponding pretext task, such as image rotation \cite{my_1gidaris2018unsupervised}. The current mainstream SSL approach is based on the instance discrimination task \cite{47wu2018unsupervised} under the contrastive learning framework. Briefly, each image in the dataset is treated as a separate semantic class. In feature space, augmented views of the same image are pulled closer and views between other images are pushed away by the noise contrastive estimation (NCE) loss \cite{my_8gutmann2010noise}. Meanwhile, there are some milestone works to continuously improve the methods under contrastive learning. For example, MoCo \cite{24he2020momentum} introduced momentum encoders and queues to address the dilemma of memory consumption and untimely updates of data features. SimCLR \cite{6chen2020simple} increased the difficulty of the self-supervised pre-training task by applying complex data augmentation and additional non-linear projectors.

The problem of false negative samples has severely impeded the ongoing development of contrastive self-supervised learning. False negative samples are defined as samples within the negative sample set that exhibit similar semantics or categories to the positive sample. Some works have attempted to address the false negative sample problem by introducing nearest neighbors (NN). For example, NNCLR \cite{my_9dwibedi2021little} searches for the nearest neighbors of the query sample in its imported support set and performs NCE loss with the positive sample. MSF \cite{my_6koohpayegani2021mean} enriches the semantic information of the positive sample by searching for the top-K neighbors in the Memory Bank \cite{47wu2018unsupervised} and performing the Mean Squared Error loss with the query sample. CMSF \cite{cmsf_2022} improves the semantic diversity of neighbor samples with an additional Memory Bank. SNCLR \cite{snclr_2023} leverages cross-attention scores to distinguish the contribution of different neighbor samples to the model. However, these methods often require a predefined number of neighboring samples, and determining this number in advance can be challenging.

In this study, we are interested in improving the reliability and coverage of models to identify false negative samples. Inspired by MSF, we propose a new simple Self-Supervised Learning with Multiple Sample Views and Queues (MSVQ). Our approach employs two complementary and symmetrical methods within the teacher networks to generate three distinct soft labels for the student network. Firstly, we create multiple augmented views of the positive sample within the teacher networks and perform consistent similarity distillation with negative samples from the same queue. Secondly, we introduce two separate queues into the model to generate diverse semantic features for the negative samples within these queues. This is accomplished by utilizing two momentum encoders with different update coefficients.

Our main contributions are summarized as follows:
\begin{itemize}[itemsep=4pt,topsep=0pt,parsep=0pt]
    \item More augmented views of the positive sample. In the teacher networks, we apply weak data augmentation to the positive sample multiple times to enhance feature diversity. It is then distilled for consistency similarity with the negative samples in the queues to improve the reliability of the model in identifying false negative samples.
    \item Feature diversity of negative samples. We form the variability of features on the same negative samples by leveraging encoders with different momentum coefficients. Our intuition is that the embedding diversity of negative samples can reduce the risk of omission for the model to identify false negative samples in feature space.
\end{itemize}
\section{Related Work}\label{2}
\subsection{Self-supervised learning}\label{21}
Self-supervised learning is an approach for learning generic semantic features from data by solving pretext tasks using large amounts of unlabeled data. In the early SSL, common pretext tasks included rotations \cite{35lee2020_sla, my_1gidaris2018unsupervised}, grayscale coloring \cite{50zhang2016colorful}, or cropping \cite{39noroozi2016unsupervised, 16doersch2015unsupervised}. However, this approach may result in semantic features containing noisy information related to specific pretext tasks, which can hinder generalization \cite{my_1gidaris2018unsupervised, my_2kolesnikov2019revisiting}. 

In recent years, contrastive learning methods based on instance discrimination tasks \cite{32lai2019contrastive, 40tian2020contrastive, 6chen2020simple, 24he2020momentum, 41tian2020makes, 29huang2022self, 27hjelm2018learning, 30kalantidis2020hard} have achieved rapid development in SSL. The core idea is that the semantic features of augmented views generated from the same image should be invariant. There are also some classical works under this framework. For instance, MoCo is designed to mitigate rapid memory consumption caused by too large batchsize, and performance decreases due to data features in the memory bank not being updated in time. This is achieved by employing the momentum encoder and the queue. SimCLR enhances the generalization of data features by introducing complex data augmentation and additional nonlinear projectors. Un-Mix \cite{unmix_2022} uses the idea of mix-up \cite{mixup_2018} to make the model predict less confidently. Meanwhile, several recent studies have demonstrated the efficiency of contrastive learning by leveraging feature similarities among positive samples. For example, BYOL \cite{23grill2020bootstrap} prevents model collapse without using negative samples based on the introduction of predictor and Mean Squared Error (MSE) loss. SimSiam \cite{9chen2021exploring} demonstrates that a simple siamese framework with stop-grad can achieve favorable properties on its own without requiring additional components.
\begin{figure}[!h]
\centering
        \includegraphics[width=90mm]{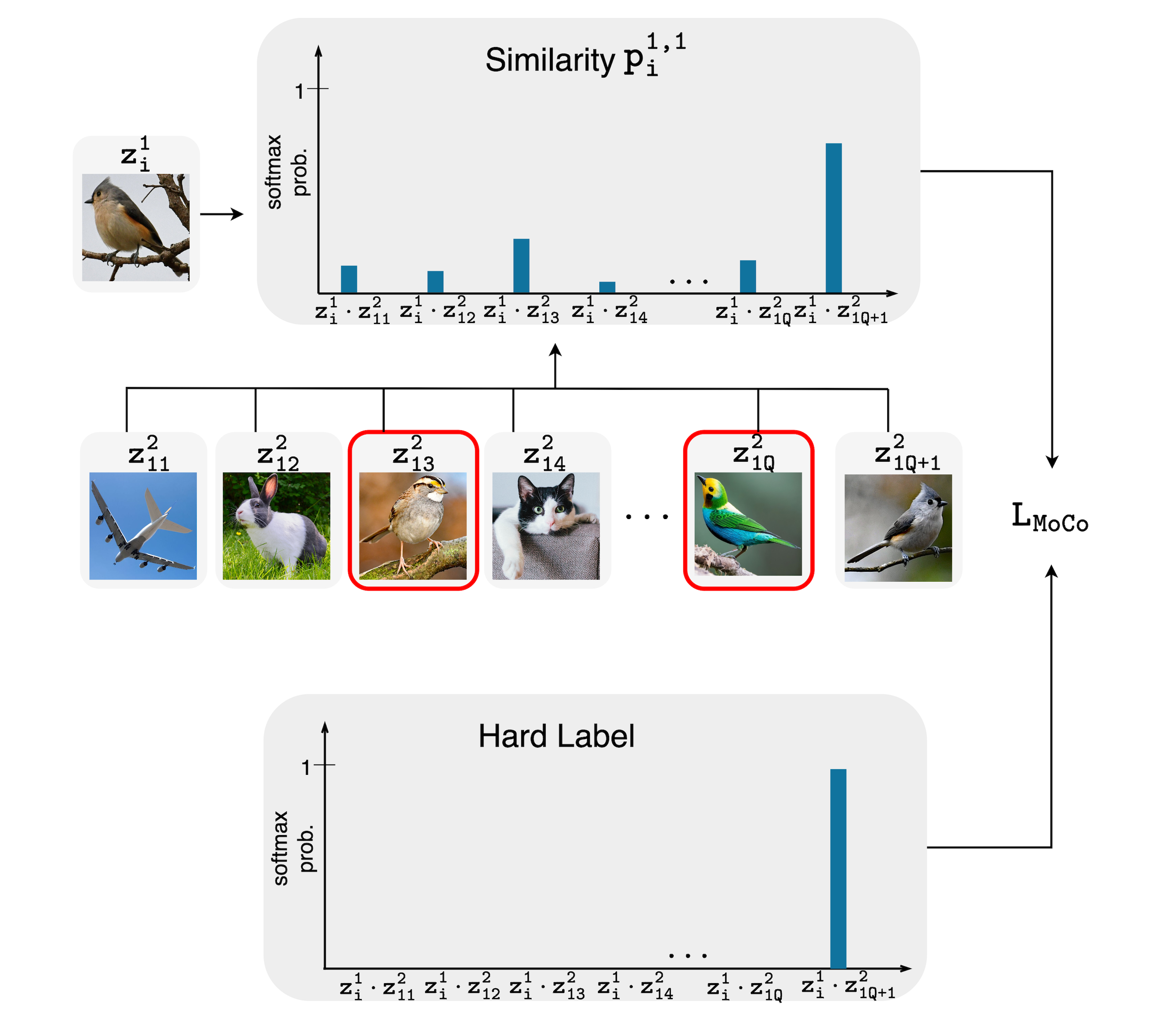}
	\centering
     \caption{Illustration of MoCo. While the similarity distribution $p_i^{1,1}$ can search for negative samples (the samples marked by red blocks, i.e., false negative samples) in the queue that are semantically similar to the query sample $z^1_i$. But an artificial one-hot label ignores the semantic relationship between them.}
     \label{figure: illustration_moco}
 \vspace{-1.0em}
\end{figure}
\subsection{Knowledge distillation}\label{22}
Knowledge distillation \cite{isd22hinton2015distilling, isd3ba2014deep} is a method where a model (teacher network) that has learned rich semantic features is used to transfer its knowledge to another model (student network). While in most cases, the teacher network has a more complex model structure compared to the student network. In some work, the teacher network and the student network can also be the same model structure \cite{isd5bagherinezhad2018label, isd16furlanello2018born}.

Recent contrastive self-supervised learning can be seen as a structure that contains a student network and a teacher network. For example, in MoCo, the student network corresponds to an encoder with stochastic gradient descent (SGD) update and the teacher network corresponds to an encoder with momentum update. Compared to previous work \cite{my_4tejankar2021isd, 21fang2021seed}, the student network in the proposed MSVQ can learn rich semantic features from multiple teacher networks with different knowledge. The student network has a more flexible ability to identify false negative samples in the dataset. MOKD \cite{my_10song2023multi} is the closest related work to our approach that utilizes multiple teacher networks to teach a student network. However, our approach differs in that we use different momentum update coefficients to construct teacher networks, rather than introducing multiple heterogeneous models (e.g., a combination of ResNet \cite{25he2016deep} and ViT \cite{my_11dosovitskiy2021an}).
\subsection{Consistency regularization}\label{23}
Consistency regularization is an approach to make the output of a model consistent or similar under small perturbations in the input \cite{isd29miyato2018virtual} or model parameters \cite{isd41tarvainen2017mean}. It allows the model to learn the most possible diversity of semantic features of the input data. Inspired by this idea, some works have been applied to contrastive self-supervised learning to solve the false negative sample problem. For example, MSF introduces top-K neighbors of the positive sample and minimizes the distance between the query sample and its nearest neighbors. This helps the network learn the semantic diversity of the query sample. On the SimCLR-based framework, NNCLR attempts to search for nearest neighbors in the support set and performs NCE loss with the positive sample. However, the performance of these methods is sensitive to the number of nearest neighbors K. Meanwhile, due to the random initialization, it is meaningless for the model to find the nearest neighbors in the early training stage.

CO2 \cite{isd45wei2020co2} is based on MoCo by adding a consistency regularization term to distill various augmented views of the same image with negative samples. However, ReSSL \cite{my_3zheng2021ressl} shows that the regularization term itself can learn meaningful information when the appropriate temperature parameter and data augmentation are utilized in the teacher network. In addition, SCE \cite{my_7denize2023similarity} incorporates hard labels into ReSSL to enhance the discriminative ability of the model. Our approach utilizes multiple augmented views of the positive sample along with consistent similarity distillation of negative samples within the queues. Inspired by the concept of symmetry, we incorporate two momentum encoders to generate distinct semantic features for the same negative samples. Both of these complementary approaches are employed to jointly improve the ability of the model to identify false negative samples.
\begin{figure}[!h]
	\centering
    \includegraphics[width=90mm]{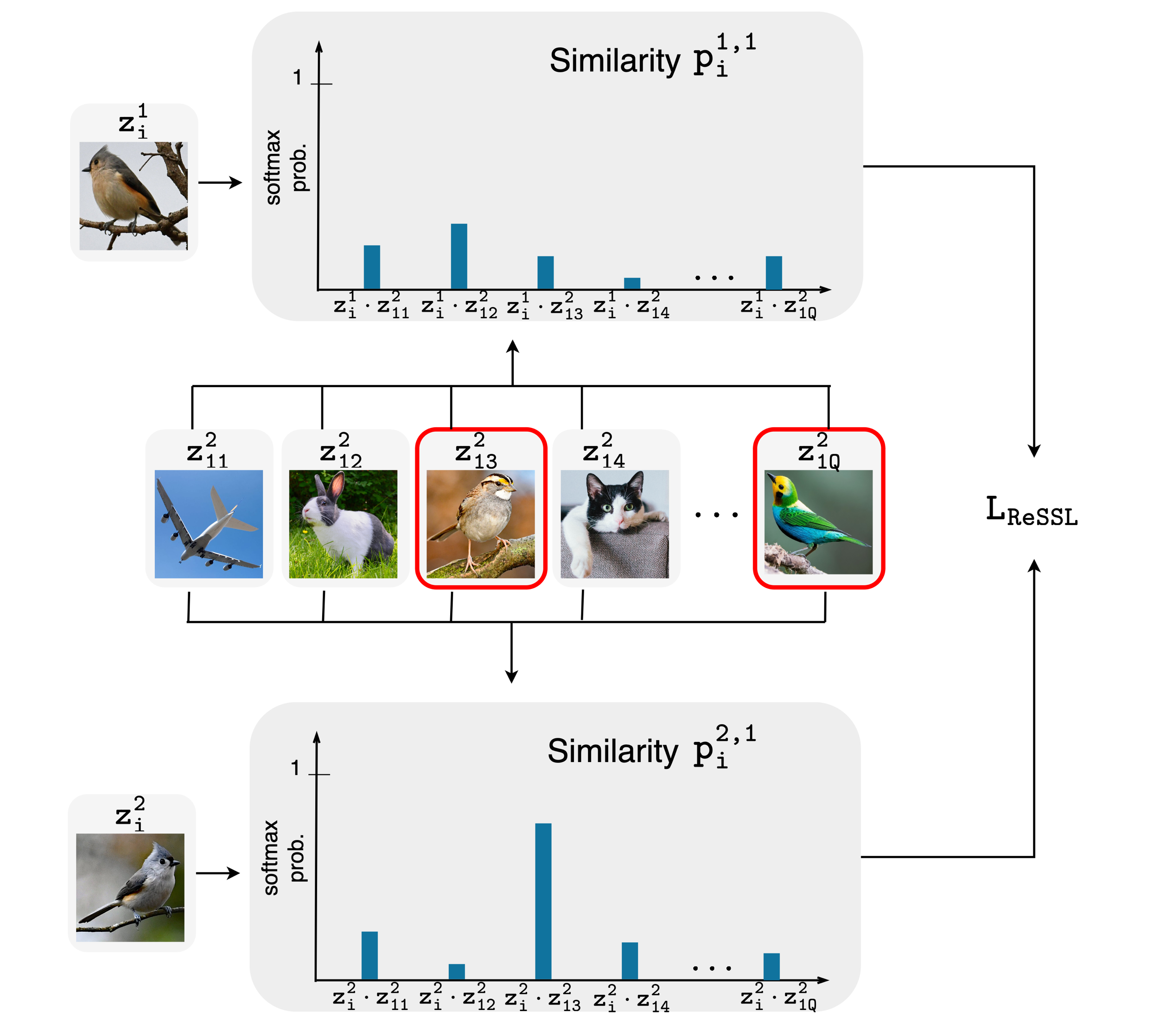}
	\centering
     \caption{Illustration of ReSSL. We can correspond the upper and lower branches to the student network and the teacher network, respectively. Since the teacher network uses an appropriate temperature parameter and weak data augmentation. $p_i^{2,1}$ highlights the important instance relationships and filters out some trivial connections. Nevertheless, depending on a single teacher network to accurately and consistently represent the similarity between instance samples is challenging.
     }
     \label{figure: illustration_ressl}
     \vspace{-1.0em}
\end{figure}
\begin{figure*}[!h]
	\centering
    \includegraphics[width=140mm]{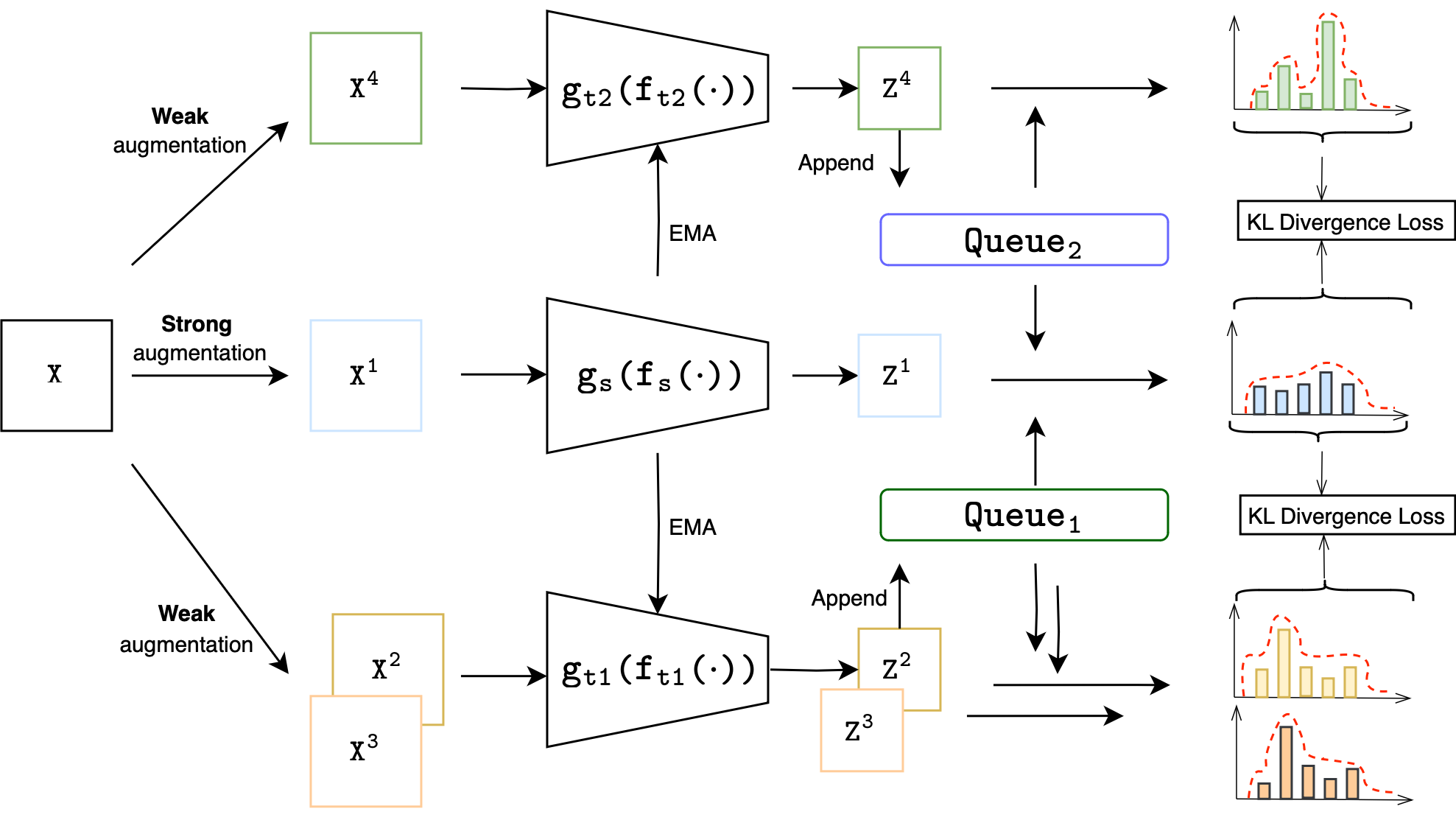}
	\centering
     \caption{The overall framework of MSVQ. We can consider the relationship distributions generated by the two teacher networks as three distinct soft labels, which serve as guidance for the student network in classifying the negative samples within the queues.}
     \label{figure: illustration_msvq}
     \vspace{-1.0em}
\end{figure*}
\section{Methodology}\label{3}
In this section, first, we briefly review our baselines: MoCo \cite{24he2020momentum, 8chen2020improved} and ReSSL \cite{my_3zheng2021ressl}. Then we introduce our proposed MSVQ framework. Meanwhile, the algorithm and implementation details of MSVQ will also be explained.
\subsection{Previous contrastive self-supervised learning and relational learning}\label{31}
Let $X \in R^{N\times H\times W\times C}$ be $N$ training samples with height $H$, width $W$, and number of channels $C$. The queues $Queue_1 = {\{z_{1j}^2\}}_1^Q$ and $Queue_2 = {\{z_{2j}^4\}}_1^Q$ each contain a set of $Q$ random embeddings of other samples. Meanwhile, $f\_(\cdot)$ is a backbone network (e.g., ResNet \cite{25he2016deep}) and $g\_(\cdot)$ is a nonlinear projector. First, we generate two different views by data augmentation ($T(\cdot)$) on the training samples: $X^1 = T_1(X)$ and $X^2 = T_2(X)$. Next, we input these views into the student network and the teacher network to obtain the corresponding embedding features $Z^1 = g_s(f_s(X^1))$ and $Z^2 = g_{t1}(f_{t1}(X^2))$, respectively. In the case of MoCo (as depicted in Fig.~\ref{figure: illustration_moco}), it employs the NCE loss, which is defined as follows:
\begin{equation}
L_{MoCo} = -\frac{1}{N} {\textstyle \sum_{i=1}^{N}} log\frac{exp(sim(z^1_i, z^2_{i})/\tau)}{{\textstyle \sum_{j=1}^{Q+1}exp(sim(z^1_i, z^2_{1j})/\tau)}} \label{eq:moco}
\end{equation}
where $z^2_{1Q+1} \triangleq z^2_i$ and $\tau$ is the temperature parameter. Meanwhile, $sim(u, v)$ is the similarity measure of the two feature embeddings, such as cosine similarity:
\begin{equation}
sim(u, v) = \frac{u^Tv}{||u||_2||v||_2}\label{eq:cos}
\end{equation}
In the context of ReSSL (as illustrated in Fig.~\ref{figure: illustration_ressl}), the method utilizes distributions $p_{ij}^{1,1}$ and $p_{ij}^{2,1}$ to represent the similarity relationship among the instance samples:
\begin{equation}
p^{1,1}_{ij} = \frac{exp(sim(z^1_i, z^2_{1j})/\tau_s)}{ {\textstyle \sum_{k=1}^{Q}exp(sim(z^1_i, z^2_{1k})/\tau_s)}},\space j = 1, 2, ..., Q 
\label{eq:relationship_distribution_1_ressl}
\end{equation}
\begin{equation}
p^{2,1}_{ij} = \frac{exp(sim(z^2_i, z^2_{1j})/\tau_t)}{ {\textstyle \sum_{k=1}^{Q}exp(sim(z^2_i, z^2_{1k})/\tau_t)}},\space j = 1, 2, ..., Q 
\label{eq:relationship_distribution_2_ressl}
\end{equation}
where $\tau_s$ and $\tau_t$ represent the temperature hyperparameters used in calculating the relationship distributions within the student network and the teacher network, respectively. The loss function of ReSSL is the KL divergence of $P^{2,1} $ and $P^{1,1}$:
\begin{equation}
L_{ReSSL} = KL(P^{2,1}||P^{1,1})\label{eq:KL_ressl}
\end{equation}

Both MoCo and ReSSL employ a similar approach to update their teacher networks, as described by the following formula:
\begin{equation}
f_{t1} = m_1f_{t1} + (1-m_1)f_{s}, \space{g_{t1} = m_1g_{t1} + (1-m_1)g_s} \label{eq:ema_ressl}
\end{equation}
where $m_1$ indicates the momentum update coefficient. Since the teacher network in ReSSL is not updated directly by the loss function, its loss function can simply use cross entropy instead of KL divergence. We also employ the momentum update mechanism, queue storage for negative samples, and KL divergence to capture inter-sample relationships.
\subsection{Self-supervised learning with multiple sample views and queues}\label{32}
In this work, we propose two symmetric and complementary ways to improve the reliability and coverage of the model to identify false negative samples.
\subsubsection{Multiple sample views}\label{321}
Inspired by some work \cite{my_6koohpayegani2021mean, my_9dwibedi2021little, my_3zheng2021ressl}, we incorporate multiple augmented views of the positive sample into the teacher network to comprehensively represent its semantics. This approach aids in the precise identification of false negative samples within the queue $Queue_1$. Specifically, the training samples denoted as $X$ undergo data augmentation via $T(\cdot)$, resulting in $X^3 = T_3(X)$. Subsequently, it is projected using $g_{t1}(f_{t1}(\cdot))$ to obtain $Z^3$. Finally, we calculate the similarity between $Z^3$ and the negative samples in $Queue_1$, applying the softmax function to derive a distribution that represents the relationships among these samples:
\begin{equation}
p^{3,1}_{ij} = \frac{exp(sim(z^3_i, z^2_{1j})/\tau_t)}{ {\textstyle \sum_{k=1}^{Q}exp(sim(z^3_i, z^2_{1k})/\tau_t)}},\space j = 1, 2, ..., Q 
\label{eq:relationship_distribution_msv_1}
\end{equation}
$P^{3,1}$ serves as an extra soft label to guide the student network. This guidance involves calculating the KL divergence between $P^{3,1}$ and $P^{1,1}$. To simplify, we average its loss function with ReSSL:
\begin{equation}
L_{MSV} = \frac{1}{2}(KL(P^{2,1}||P^{1,1}) + KL(P^{3,1}||P^{1,1}))\label{eq:KL_msv}
\end{equation}
The update rule of the teacher network in Multiple Sample Views (MSV) is consistent with that of ReSSL.
\subsubsection{Multiple queues}\label{322}
MSV can be viewed as a variant of previous work \cite{my_6koohpayegani2021mean, my_9dwibedi2021little}, with the notable distinction that MSV enhances the semantics of the positive sample through multiple data augmentations rather than identifying its top-K neighbors. From a symmetric perspective, we also focus on the aspect of negative samples. Our intuition is that the diversity of both is complementary and can help achieve a more comprehensive and robust representation of the underlying data distribution.

To be specific, a teacher network $g_{t2}(f_{t2}(\cdot))$ with a momentum coefficient of $m_2$ is employed. The training images $X$ are independently subjected to data augmentation using $T(\cdot)$ to obtain $X^4 = T_4(X)$, and then they are projected using $g_{t2}(f_{t2}(\cdot))$ to yield $Z^4 = {\{z_i^4\}}_i^{N}$. The following relationship distribution indicates the similarity between instances:
\begin{equation}
p^{1,2}_{ij} = \frac{exp(sim(z^1_i, z^4_{2j})/\tau_s)}{ {\textstyle \sum_{k=1}^{Q}exp(sim(z^1_i, z^4_{2k})/\tau_s)}},\space j = 1, 2, ..., Q 
\label{eq:relationship_distribution_1_ressl}
\end{equation}
\begin{equation}
p^{4,2}_{ij} = \frac{exp(sim(z^4_i, z^4_{2j})/\tau_t)}{ {\textstyle \sum_{k=1}^{Q}exp(sim(z^4_i, z^4_{2k})/\tau_t)}},\space j = 1, 2, ..., Q 
\label{eq:relationship_distribution_mq_1}
\end{equation}

This part is the KL divergence of $P^{4,2} $ and $P^{1,2}$. For simplicity, we also just take the average value of its loss function with ReSSL:
\begin{equation}
L_{MQ} = \frac{1}{2}(KL(P^{2,1}||P^{1,1}) + KL(P^{4,2}||P^{1,2}))\label{eq:KL_mq}
\end{equation}

The Multiple Queues (MQ) contains two teacher networks with different update coefficients. The first teacher network is updated in the identical way as ReSSL, while the second teacher network uses the following update mechanism:
\begin{equation}
f_{t2} = m_{2}f_{t2} + (1-m_2)f_{s}, \space{g_{t2} = m_2g_{t2} + (1-m_2)g_{s}} \label{eq:ema_mq_2}
\end{equation}

\subsubsection{Multiple sample views and queues}\label{323}
As illustrated in Fig.~\ref{figure: illustration_msvq}, a simple organic merging of these two symmetric ways yields our proposed method. We optimize the student network only by minimizing the following loss:
\begin{small}  
\begin{equation}
L_{MSVQ} = \frac{1}{3}(KL(P^{2,1}||P^{1,1}) +KL(P^{3,1}||P^{1,1}) + KL(P^{4,2}||P^{1,2}))\label{eq:KL_msvq}
\end{equation}
\end{small}

This method combines the strengths of both approaches to generate three distinct soft labels, $P^{2,1}$, $P^{3,1}$, and $P^{4,2}$, aiming to mitigate the under-detection of false negative samples by the model. These labels effectively transfer knowledge from multiple teacher networks to the student network. Further analysis and discussion of these three soft labels are presented in Sec.~\ref{46}.

Notably, $z_{tj}^\_$ represents the semantic embedding of negative sample $j$ in $Queue_t$, $p^{\_}_{ij}$ signifies the inter-instance similarity between the positive sample $z_i^\_$ and the negative sample $z_{tj}^\_$. We have included a comprehensive list of important notations in the MSVQ framework along with their specific meanings in Table~\ref{table: symbols}. The procedure of MSVQ is summarized in Algorithm~\ref{algo:algo-msvq}.
\subsubsection{Sharper distribution and friendly data augmentation}\label{324}
To effectively emphasize significant false negative samples within the teacher networks and simultaneously filter out potential noisy relationships in the queues, our model ensures that $\tau_t < \tau_s$. We conduct an analysis of the impact of different $\tau_t$ values on MSVQ in Sec.~\ref{451}.

To reduce the impact of aggressive data augmentation \cite{6chen2020simple} in the teacher networks, we employ a weaker data augmentation scheme \cite{my_6koohpayegani2021mean, my_7denize2023similarity, my_3zheng2021ressl} to generate more suitable soft labels. We also conduct an analysis of the impact of various data augmentation ways on MSVQ in Sec.~\ref{452}.
\begin{table}[htb]
\centering
\Large
\caption{Notations in the MSVQ Framework.}
\resizebox{\linewidth}{!}
{
\begin{tabularx}{\textwidth}{p{0.2\textwidth}X}
\hline
Notation & Meaning \\ \hline 
$X = {\{x_i\}}_1^N$ & Comprising N training samples from the current batch, where each element is regarded as a positive sample. \vspace*{0.4\baselineskip} \\
$Queue_1 = {\{z_{1j}^2\}}_1^Q$  ($Queue_2 = {\{z_{2j}^4\}}_1^Q$) & Comprising Q negative sample features, with these features derived from the most recent previous batches of $Z^2$ ($Z^4$). The update method follows a first-in-first-out (FIFO) approach. \vspace*{0.4\baselineskip} \\
$T(\cdot)$ & The data augmentation distribution is employed to generate various augmented views. Specifically, $T_1(\cdot)$ is utilized to apply strong data augmentation to the student network, while $T_i(\cdot), i\in{\{2,3,4\}}$ is employed for applying weak data augmentation to the teacher networks.\vspace*{0.4\baselineskip} \\
$Z$ & $Z^i = {\{z_j^i\}}_{j=1}^N, i\in {\{1, 2, 3, 4\}}$ denote the 128-dimensional image features obtained by projecting $X^i, i \in {\{1,2,3,4\}} $ through the corresponding network. \vspace*{0.4\baselineskip} \\
$P^{1,1}$ ($P^{1,2}$) & The similarity relationship exists between the features $Z^1$ of the positive samples and the features of the negative samples within $Queue_1$ ($Queue_2$). \vspace*{0.4\baselineskip} \\
$P^{2,1}$ ($P^{3,1}$) & The similarity relationship exists between the features $Z^2$ ($Z^3$) of the positive samples and the features of the negative samples within $Queue_1$. \vspace*{0.4\baselineskip} \\
$P^{4,2}$ & The similarity relationship exists between the features $Z^4$ of the positive samples and the features of the negative samples within $Queue_2$.\vspace*{0.4\baselineskip} \\
$m_1$ ($m_2$) & The momentum update coefficient is used to update the teacher network $g_{t1}(f_{t1}(\cdot))$ ($g_{t2}(f_{t2}(\cdot))$). To introduce variability in the semantic features of different teacher networks, we set $m_1 \neq m_2$.\vspace*{0.4\baselineskip} \\
$\tau_s$ ($\tau_t$) & Temperature hyperparameter for similarity distributions in the student network (teacher networks). \\ \hline
\end{tabularx}
}
\label{table: symbols}
\end{table}
\definecolor{commentcolor}{RGB}{110,154,155}   
\newcommand{\PyComment}[1]{\ttfamily\textcolor{commentcolor}{\# #1}}  
\newcommand{\PyCode}[1]{\ttfamily\textcolor{black}{#1}} 
\begin{algorithm}[!h]
\footnotesize 
\begin{spacing}{0.8}
\SetAlgoLined
    \PyComment{$\mathtt {F_s}$, $\mathtt {F_{t1}}$, $\mathtt {F_{t2}}$: encoder for student, $\mathtt {teacher_1}$ and $\mathtt {teacher_2}$, $\mathtt {F\_\triangleq g\_(f\_(\cdot))}$} \\
    \PyComment{$\mathtt {queue_1}$, $\mathtt {queue_2}$: two queues(CxQ)} \\
    \PyComment{$\mathtt {m_1}$, $\mathtt {m_2}$: momentum for $\mathtt {teacher_1}$ and $\mathtt {teacher_2}$} \\
    \PyComment{$\mathtt {\tau_s}$, $\mathtt {\tau_t}$: temperature for student and teacher} \\
    \PyComment{CE: CrossEntropyLoss} \\
    \vspace*{\baselineskip}
    \PyCode{$\mathtt {F_{t1}}$.params = $\mathtt {F_s}$.params} \PyComment{initialize $\mathtt {teacher_1}$} \\
    \PyCode{$\mathtt {F_{t2}}$.params = $\mathtt {F_s}$.params} \PyComment{initialize $\mathtt {teacher_2}$} \\
    \PyComment{load a minibatch X with N samples}\\
    \PyCode{for X in loader: } \\
    \Indp   
        \PyComment{random augmentation} \\
        \PyCode{$\mathtt {X^1}$, $\mathtt {X^2}$ = strong\_aug(X), weak\_aug(X) } \\
        \PyCode{$\mathtt {X^3}$, $\mathtt {X^4}$ = weak\_aug(X), weak\_aug(X)} \\
        \vspace*{\baselineskip}
        \PyCode{$\mathtt {Z^1}$, $\mathtt {Z^2}$ = $\mathtt {F_s}$.forward($\mathtt {X^1}$), $\mathtt {F_{t1}}$.forward($\mathtt {X^2}$)} \PyComment{NxC} \\
        \PyCode{$\mathtt {Z^3}$, $\mathtt {Z^4}$ = $\mathtt {F_{t1}}$.forward($\mathtt {X^3}$), $\mathtt {F_{t2}}$.forward($\mathtt {X^4}$)} \\
        \PyComment{$\mathtt {l_2-normalize}$} \\
        \PyCode{$\mathtt {Z^1}$, $\mathtt {Z^2}$, $\mathtt {Z^3}$, $\mathtt {Z^4}$ = normalize($\mathtt {Z^1}$, $\mathtt {Z^2}$, $\mathtt {Z^3}$, $\mathtt {Z^4}$, dim=1)} \\
        \vspace*{\baselineskip}
        \PyCode{$\mathtt {Z^2}$, $\mathtt {Z^3}$, $\mathtt {Z^4}$ = $\mathtt {Z^2}$.detach(), $\mathtt {Z^3}$.detach(), $\mathtt {Z^4}$.detach()} \\
        \PyComment{mm: matrix multiplication} \\
        \PyCode{$\mathtt {logits_{11}}$ = mm($\mathtt {Z^1}$, $\mathtt {queue_1}$) } \PyComment{[NxC, CxQ] -> NxQ} \\
        \PyCode{$\mathtt {logits_{12}}$ = mm($\mathtt {Z^1}$, $\mathtt {queue_2}$)} \\
        \PyCode{$\mathtt {logits_{21}}$, $\mathtt {logits_{31}}$ = mm($\mathtt {Z^2}$, $\mathtt {queue_1}$), mm($\mathtt {Z^3}$, $\mathtt {queue_1}$)} \\
        \PyCode{$\mathtt {logits_{42}}$ = mm($\mathtt {Z^4}$, $\mathtt {queue_2}$)} \\
        \vspace*{\baselineskip}
        \PyCode{$\mathtt {loss_1}$ = CE($\mathtt {logits_{11}}$/$\mathtt {\tau_s}$, softmax($\mathtt {logits_{21}}$/$\mathtt {\tau_t}$))} \\
        \PyCode{$\mathtt {loss_2}$ = CE($\mathtt {logits_{11}}$/$\mathtt {\tau_s}$, softmax($\mathtt {logits_{31}}$/$\mathtt {\tau_t}$))} \\
        \PyCode{$\mathtt {loss_3}$ = CE($\mathtt {logits_{12}}$/$\mathtt {\tau_s}$, softmax($\mathtt {logits_{42}}$/$\mathtt {\tau_t}$))} \\
        \PyCode{loss = ($\mathtt {loss_1}$+$\mathtt {loss_2}$+$\mathtt {loss_3}$)/3} \\
        \vspace*{\baselineskip}
        \PyCode{loss.backward()} \\
        \PyComment{SGD update: student} \\
        \PyCode{update($\mathtt {F_s}$.params)} \\
        \PyComment{momentum update: $\mathtt {teacher_1}$ and $\mathtt {teacher_2}$} \\
        \PyCode{$\mathtt {F_{t1}}$.params = $\mathtt {m_1}$*$\mathtt {F_{t1}}$.params+(1-$\mathtt {m_1}$)*$\mathtt {F_s}$.params} \\
        \PyCode{$\mathtt {F_{t2}}$.params = $\mathtt {m_2}$*$\mathtt {F_{t2}}$.params+(1-$\mathtt {m_2}$)*$\mathtt {F_s}$.params} \\
        \PyComment{update two queues} \\
        \PyCode{enqueue($\mathtt {queue_1}$, $\mathtt {Z^2}$)} \\
        \PyCode{enqueue($\mathtt {queue_2}$, $\mathtt {Z^4}$)} \\
        \PyCode{dequeue($\mathtt {queue_1}$)} \\
        \PyCode{dequeue($\mathtt {queue_2}$)} \\
        \vspace*{\baselineskip}
    \Indm 
\caption{PyTorch-style pseudocode for MSVQ}
\label{algo:algo-msvq}
\end{spacing}
\end{algorithm}
\begin{table}[!h]
\centering
\setlength{\belowcaptionskip}{6 pt}
\small
\caption{Parameters of the experiment.}
\resizebox{\linewidth}{!}
{
\begin{tabular}{lcccc}
\hline
                         & CIFAR-10 & CIFAR-100 & STL-10       & Tiny ImageNet \\ \hline
\textbf{Pre-training}        &          &           &              &               \\ \hline
Epoch                    & 200      & 200       & 200          & 200           \\
Batch size               & 256      & 256       & 256          & 256           \\
Warm up epoch            & 5        & 5         & 5            & 5             \\
Base learning rate       & 0.06     & 0.06      & 0.06         & 0.06          \\
($m_1$,~$m_2$)      & (0.99,~0.95) & (0.99,~0.93)  & (0.996,~0.99) & (0.996,~0.99)  \\
($\tau_s$,~$\tau_t$)  & (0.1,~0.04)  & (0.1,~0.03)   & (0.1,~0.04)   & (0.1,~0.04)    \\
Queue size               & 4096     & 4096      & 16384        & 16384         \\
Weight decay             & 5e-4     & 5e-4      & 5e-4         & 5e-4          \\ 
Cropped and Resized      & $32 \times 32$ & $32 \times 32$  & $64 \times 64$ & $64 \times 64$ \\ \hline
\textbf{Fine-tuning} &          &           &              &               \\ \hline
Epoch                    & 100      & 100       & 100          & 100           \\
Batch size               & 256      & 256       & 256          & 256           \\
Base learning rate       & 1        & 1         & 1            & 1             \\
Weight decay             & 0        & 0         & 0            & 0             \\ \hline
\end{tabular}
}
\label{table: experiment_parameters}
\vspace{-1.0em}
\end{table}
\begin{table}[!h]
\centering
\setlength{\belowcaptionskip}{6 pt}
\small
\caption{Data augmentation of the experiment. 'Resized Crops' specifies the lower and upper bounds of scale for cropping a random region based on the area of the original image. For the other data augmentations, they indicate the probability (denoted as '$\rho$') of being randomly applied. For instance, in the Strong category, the probability that "Horizontal Flip" is applied is $\rho=0.5$.}
\resizebox{\linewidth}{!}
{
\begin{tabular}{lccccc}
\hline
       & Resized Crops & Horizontal Flip & Color Jitter & GrayScale & Gaussian Blur \\ \hline
Strong & (0.2, 1.0)    & 0.5             & 0.8    & 0.2 & 0.5     \\        
Weak   & (0.2, 1.0)    & 0.9             &     &  &      \\    
\hline
\end{tabular}
}
\label{table: data_augmentation}
\vspace{-1.0em}
\end{table}
\section{Experiments and Results}\label{4}
In this section, we will compare and analyze the proposed MSVQ with previous classical algorithms on four benchmark image datasets.
\subsection{Datasets and device performance}\label{41}
Most SSL methods are typically evaluated using an ImageNet-1K dataset \cite{14deng2009imagenet} containing almost 1.3M images. However, due to hardware limitations, implementing this evaluation can be challenging for most research labs. We evaluate the proposed method MSVQ with some classical contrastive self-supervised methods on four datasets.
\begin{itemize}
    \item CIFAR-10 and CIFAR-100 \cite{31krizhevsky2009learning}: Both datasets comprise 60,000 color images, consisting of 50,000 training images and 10,000 test images, all with a resolution of 32x32 pixels. However, they differ in the number of classes they contain: the CIFAR-10 dataset comprises 10 classes, while the CIFAR-100 dataset comprises 100 semantic classes.

    \item STL-10 \cite{11coates2011analysis} and Tiny ImageNet \cite{33le2015tiny}: For the STL-10 dataset, the training set comprises 100,000 unlabeled color images and 5,000 labeled color images. Additionally, the test set includes 8,000 labeled images. All of these images share a common resolution of 96x96 pixels, and the dataset encompasses a total of 10 categories. Tiny ImageNet comprises 120,000 images distributed across 200 classes. These images are resized to a dimension of 64×64 pixels. Specifically, each class includes 500 training images, along with 50 validation images and 50 test images.
\end{itemize}

A consistent hardware setup (1 Nvidia GTX 3090 GPU) was used for all algorithm experiments in this study. All algorithms are initially pre-trained using the training set. During the evaluation phase, we evaluate them using the test set, except in the case of Tiny ImageNet, where the validation set is used.
\subsection{Pre-training}\label{42}
In all datasets, we use ResNet18 \cite{25he2016deep} as the backbone network $f(\cdot)$. Meanwhile, a nonlinear projector $g(\cdot)$ is added following the backbone network. All projectors within both the student network and teacher networks are composed of two fully-connected (FC) layers together with a linear rectification function (ReLU) layer between them, where the first FC layer is of size [512, 2048], and the second is of size [2048, 128].

To ensure a fair comparison, certain hyperparameters and data augmentations \cite{my_5feng2022adaptive} in MSVQ were aligned with those in ReSSL. Regarding the model parameters, we employed the SGD optimizer with a momentum value of 0.9 and a weight decay of 5e-4 for pre-training the model over 200 epochs. In terms of data augmentation, strong data augmentation was applied to the student network, while weak data augmentation was employed for the teacher networks. Additional details can be found in Table~\ref{table: experiment_parameters} and Table~\ref{table: data_augmentation}, respectively.

\subsection{Fine-tuning}\label{43}
After the pre-training phase, we employed the widely adopted linear evaluation protocol to assess our model. In this protocol, we initially discarded the pre-trained nonlinear projector and fixed all parameters of the backbone network. Afterward, we added a linear classifier to the backbone of the student network with dimensions [512, \textit{cla}], where \textit{cla} represents the number of categories in the dataset. This linear classifier was used for the linear evaluation, enabling the model to perform classification based on the learned features. Finally, we fine-tuned the classifier for 100 epochs using the SGD optimizer with a momentum of 0.9 and weight decay of 0. Further details of these parameters can be found in Table~\ref{table: experiment_parameters}.
\begin{figure*}[htb]
	\centering
	\subfigure[CIFAR-10]{
		\begin{minipage}[t]{0.32\linewidth}
			\centering
			\includegraphics[width=\textwidth]{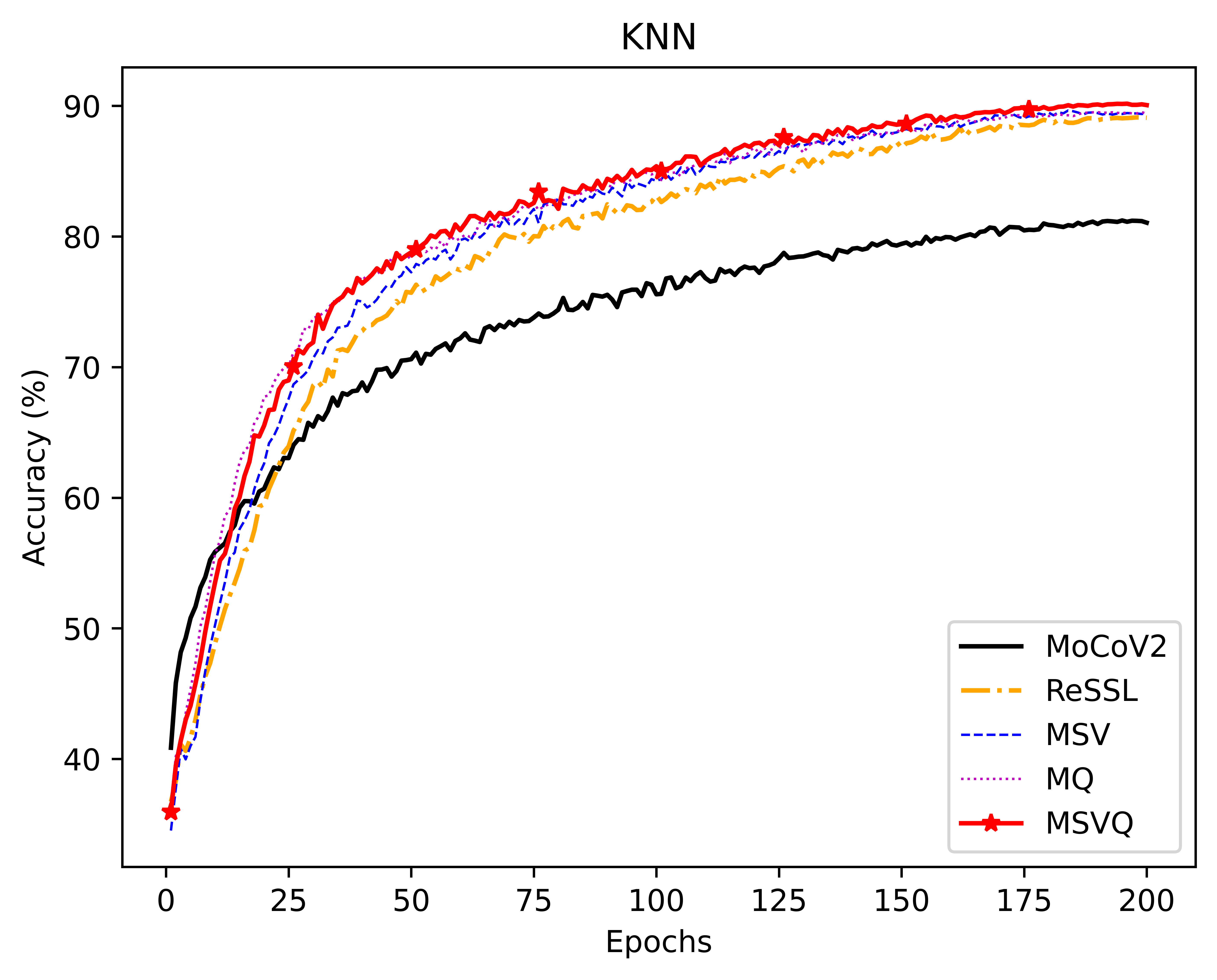}
		\end{minipage}
	}%
	\subfigure[CIFAR-100]{
		\begin{minipage}[t]{0.32\linewidth}
			\centering
			\includegraphics[width=\textwidth]{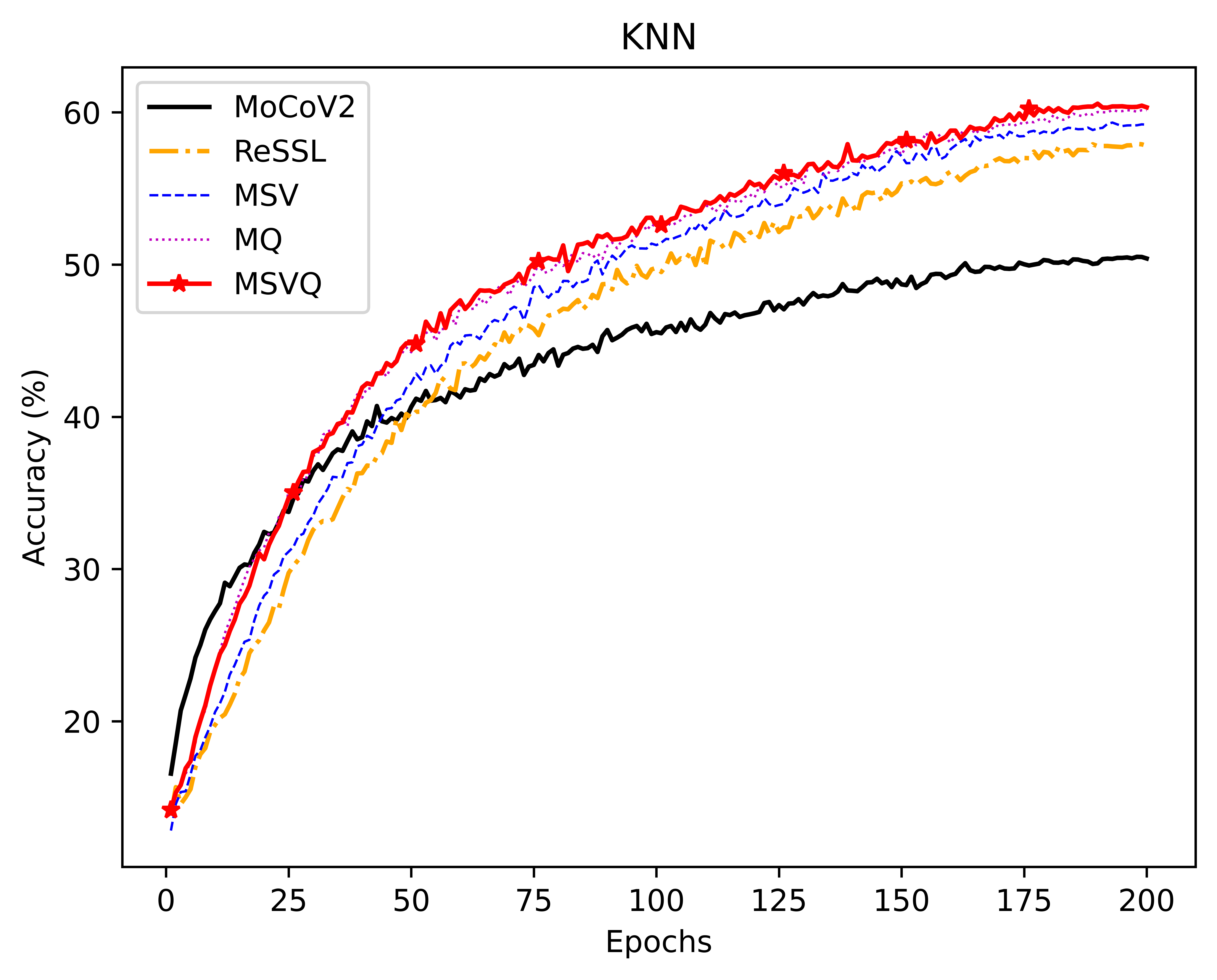}
		\end{minipage}
	}%
	\centering
	\caption{KNN online evaluation accuracy curve.}
	\label{figure:knn}
\end{figure*}
\subsection{Main results}\label{44}
\subsubsection{Linear evaluation protocol}\label{441}
In Table~\ref{table: linear_eval}, the results of other methods are copied from \cite{my_3zheng2021ressl} for best results. For a fair comparison, we have also reproduced some recent work and marked it with *. It is clear that MSVQ outperforms the other classical methods on most benchmarks. Noticeably, MSVQ also has a significantly better performance compared to the MSV and the MQ trained alone. This indicates that the MSV and MQ methods learn complementary semantic features of images. Meanwhile, our method requires only a slight additional overhead without multiple backpropagations.
\begin{table}[ht]
\centering
\setlength{\belowcaptionskip}{6 pt}
\small
\caption{Linear evaluation results. The optimal results are shown in bold, and the suboptimal results are underlined. Results marked with $^*$ were reproduced from the official code. $\dagger$: Similar to the SCE \cite{my_7denize2023similarity} network framework, batch normalization \cite{bn_2015} is incorporated after the first FC layer of the projector within MSVQ.}
\resizebox{\linewidth}{!}
{
\begin{tabular}{lccccc}
\hline
Method     & BackProp & CIFAR-10       & CIFAR-100      & STL-10         & Tiny ImageNet                \\ \hline
Supervised & -        & 94.22          & 74.66          & 82.55          & 59.26                        \\ \hline
SimCLR \cite{6chen2020simple}     & 2x      & 84.92          & 59.28          & 85.48          & 44.38                        \\
BYOL \cite{23grill2020bootstrap}       & 2x      & 85.82          & 57.75          & 87.45          & 42.70                         \\
SimSiam \cite{9chen2021exploring}    & 2x      & 88.51          & 60.00          & 87.47          & 37.04                        \\
MoCoV2 \cite{8chen2020improved}    & 1x      & 86.18          & 59.51          & 85.88          & 43.36                        \\
ReSSL \cite{my_3zheng2021ressl}      & 1x       & 90.20          & 63.79          & 88.25          & 46.60                        \\
ReSSL$^*$ \cite{my_3zheng2021ressl}   & 1x       & 90.22          & 64.22          & 87.64          & 45.61                        \\
CMSF$^*$ \cite{cmsf_2022}   & 2x       & {\ul91.00}          & 62.37          & 88.21          & 44.50                        \\
Un-Mix$^*$ \cite{unmix_2022} & 2x & 90.20 & 64.42 & 89.76 & 45.20\\
SNCLR$^*$ \cite{snclr_2023}  & 2x       & 88.50          & 62.40          & 88.24          & 46.14                        \\
SCE \cite{my_7denize2023similarity} & 2x & 90.34          & {\ul65.45}          & {\ul89.94}          &  \textbf{51.90}          \\
MSV(Ours)  & 1x       & 90.92    & 65.02          & 89.35          &  46.68 \\
MQ(Ours)   & 1x       & 90.91          & 65.15    & 89.58    &46.32 \\
\textbf{MSVQ(Ours)} & 1x       & \textbf{91.46} & \textbf{66.44} & \textbf{90.41} & {\ul48.09}               \\ \hline\hline
SCE$^*$ \cite{my_7denize2023similarity} & 2x &    90.03       & 65.41          & 90.06          &  48.11          \\
\textbf{MSVQ$\dagger$(Ours)} & 1x & 91.28  & 65.82 & 89.71 & 49.51 \\ \hline
\end{tabular}
}
 \label{table: linear_eval}
 \vspace{-1.0em}
\end{table}
\subsubsection{Learning efficiency analysis}\label{442}
To reduce the influence of downstream task hyperparameters on the model and enhance evaluation efficiency, we also employ K Nearest Neighbors (KNN) classification to assess the pre-trained off-the-shelf features, with K set to 200 \cite{47wu2018unsupervised}. The online KNN classification results in Table~\ref{table:knn_evaluation} reflect that MSVQ can learn rich semantic features in the pre-training stage. In Fig.~\ref{figure:knn}, the KNN (K=200) classification accuracy curves show that our method has reliable learning efficiency compared to ReSSL and MoCoV2.
\begin{table}[!h]
\vspace{-1.0em}
\centering
\setlength{\belowcaptionskip}{6 pt}
\small
\caption{The online evaluation accuracy using KNN with K set to 200. The best results are shown in bold and the suboptimal results are underlined. $^*$ denotes results that were reproduced from the official code. $\dagger$: Similar to the SCE \cite{my_7denize2023similarity} network framework, batch normalization is added after the first FC layer of the projector within MSVQ.}
\resizebox{\linewidth}{!}{
\begin{tabular}{lcccc} \hline
Method     & CIFAR-10       & CIFAR-100      & STL-10         & Tiny ImageNet  \\ \hline
ReSSL$^*$ \cite{my_3zheng2021ressl}   & 89.11          & 57.93          & 84.18          & 37.19          \\
CMSF$^*$ \cite{cmsf_2022}   & 89.30          & 55.57          & 84.11          & 36.79          \\
Un-Mix$^*$ \cite{unmix_2022} & 88.38 & 59.52 & 85.18 & 38.81\\
SNCLR$^*$ \cite{snclr_2023}   & 86.15          & 55.07          & 82.18          & 36.86          \\ 
MSV(Ours)  & {\ul 89.64}    & 59.33          & 85.58          & 38.78          \\
MQ(Ours)   & 89.51  & {\ul 60.16}    & {\ul 85.69}  &{\ul 38.84} \\ 
\textbf{MSVQ(Ours)} & \textbf{90.16} & \textbf{60.66} & \textbf{86.51} & \textbf{40.26} \\ \hline\hline
SCE$^*$ \cite{my_7denize2023similarity} & 88.41          & 59.30          & 85.45          &  40.17          \\ 
\textbf{MSVQ$\dagger$(Ours)} & 90.23  & 61.63 & 86.56 & 41.81 \\ \hline

\end{tabular}}
\label{table:knn_evaluation}
\vspace{-1.0em}
\end{table}
\subsection{Ablation studies}\label{45}
\subsubsection{Sharper distribution}\label{451}
An appropriate temperature parameter $\tau_t$ can eliminate the noisy relationship between the positive sample and the negative samples in the queues, thus providing accurate soft labels for the student network. Table~\ref{table:effect of tem_t} provides valuable insights into the influence of temperature parameters on our method. Notably, it becomes evident that extremely low or high values of these temperature parameters are suboptimal. Intuitively, temperature $\tau_t$ controls the degree of smoothing of the three labels in the MSVQ. As $\tau_t$ approaches 0, MSVQ becomes analogous to using three artificial one-hot codings. This means that each label solely focuses on the most similar false negative sample within the queue. Conversely, as $\tau_t$ approaches 0.1, the student network is constrained to acquiring only trivial knowledge from the teacher networks.
\begin{table}[!h]
\vspace{-1.0em}
\centering
\setlength{\belowcaptionskip}{6 pt}
\Large
\caption{Effect of different $\tau_t$ for MSVQ.}
\resizebox{\linewidth}{!}{
\begin{tabular}{lcccccccc}
\hline
Dataset       & $\tau_s$=0.1& $\tau_t$=0.01 & $\tau_t$=0.02 & $\tau_t$=0.03 & $\tau_t$=0.04  & $\tau_t$=0.05 & $\tau_t$=0.06 & $\tau_t$=0.07 \\ \hline
CIFAR-10      & - & 90.87         & 90.68         & 90.73         & \textbf{91.46} & 91.20          & 90.62         & 90.32         \\
CIFAR-100      & - & 65.68         & 65.33         & \textbf{66.44}     & 66.27 & 65.78          & 63.78         & 58.83         \\
STL-10      & - & 88.68         & 89.21         & 89.64         & \textbf{90.41} & 89.56          & 88.96         & 87.75         \\
Tiny ImageNet & - & 47.26         & 47.81         & 47.96         & \textbf{48.09} & 46.99         & 45.72         & 44.47         \\ \hline
\end{tabular}}
\label{table:effect of tem_t}
\vspace{-1.0em}
\end{table}
\subsubsection{Data augmentation}\label{452}
Unlike previous methods \cite{6chen2020simple} that use aggressive data augmentation to encourage the learning of semantics invariant to geometric transformations, MSVQ adopts a different approach. Within the MSVQ framework, the teacher networks employ milder, weak data augmentation techniques with the specific goal of preserving image semantics to provide appropriate soft labels for the student network. In this section, we conduct a comprehensive study to assess the performance impact of various data augmentation approaches in teacher networks. Table~\ref{table:weak_contrastive_aug} demonstrates that employing traditional data augmentation techniques for teacher networks can lead to semantic loss or distortion of the positive sample, resulting in incorrect soft labels that may mislead the student network.
\begin{table}[!h]
\vspace{-1.0em}
\centering
\setlength{\belowcaptionskip}{6 pt}
\small
\caption{Effect of weak augmentation in teacher networks for MSVQ.}
\resizebox{\linewidth}{!}{
\begin{tabular}{lcccc}
\hline
Teacher Aug & CIFAR-10       & CIFAR-100             & STL-10                & Tiny ImageNet         \\ \hline
Strong & 88.33              & 59.55 & 86.14 & 39.44 \\
Weak        & \textbf{91.46} & \textbf{66.44}        & \textbf{90.41}        & \textbf{48.09}        \\ \hline
\end{tabular}}
\label{table:weak_contrastive_aug}
\vspace{-1.0em}
\end{table}
\subsubsection{Different momentum coefficients}\label{453}
Some may question the necessity of employing distinct momentum update coefficients for the two teacher networks within the MSVQ framework. As demonstrated in Table~\ref{table: different_momentum}, when both teacher networks employ identical momentum update coefficients, a marginal reduction in performance is observed. We contend that $m_1 \ne m_2$ serves as a means to distinguish between $g_{t1}(f_{t1}(\cdot))$ and $g_{t2}(f_{t2}(\cdot))$. This differentiation enables a clear distinction between $Z^2$ (or $Z^3$) and $Z^4$, as well as among the negative samples within $Queue_1$ and $Queue_2$. Consequently, this approach maximizes the expression of semantic information within both the positive and negative samples.
\begin{table}[!h]
\vspace{-1.0em}
\centering
\setlength{\belowcaptionskip}{6 pt}
\small
\caption{Effect of different $m_2$ for MSVQ.}
\resizebox{\linewidth}{!}{
\begin{tabular}{lccccc}
\hline
Dataset       & $m_1 = 0.99 $ & $m_2 = 0.93$           & $m_2$ = 0.95           & $m_2$ = 0.97  & $m_2$ = 0.99  \\ \hline
CIFAR-10      & - & 90.93          & \textbf{91.46} & 91.08 & 90.88 \\
CIFAR-100     & - & \textbf{66.44} & 65.99          & 65.58 & 65.63 \\ \hline
Dataset       & $m_1 = 0.996$ & $m_2$ = 0.95           & $m_2$ = 0.99           & $m_2$ = 0.993 & $m_2$ = 0.996 \\ \hline
STL-10        & - & 90.00          & \textbf{90.41} & 90.10 & 90.18 \\
Tiny ImageNet & - & 47.59          & \textbf{48.09} & 47.55 & 48.05\\ \hline
\end{tabular}}
\label{table: different_momentum}
\vspace{-1.0em}
\end{table}
\subsubsection{The size of queues}\label{454}
In the context of the MSVQ framework, the size of queues directly corresponds to the number of negative samples. Table~\ref{table: Queue_size} illustrates the linear evaluation accuracy corresponding to various queue sizes. It is evident that larger queue sizes lead to a substantial improvement in performance. We hypothesize that a larger queue size augments the probability of the positive sample discovering false negative samples within the queues that align with its semantic context. Consequently, the features acquired by the model become more generalizable. However, as we further increase the queue size, we observe performance instability. We suspect that this phenomenon arises from the presence of stale features within the queues, which are not promptly replaced when the queue size is enlarged. Therefore, it is necessary for the MSVQ to strike a balance between enhancing the chances of discovering false negative samples and ensuring the timely update of features within the queues.
\begin{table}[!h]
\vspace{-1.0em}
\centering
\setlength{\belowcaptionskip}{6 pt}
\Large
\caption{Analysis of the size of queues (Q).}
\resizebox{\linewidth}{!}{
\begin{tabular}{lccccccc}
\hline
Dataset       & Q = 512 & Q = 1024 & Q = 2048 & Q = 4096 & Q = 8192 & Q = 16384 & Q = 32768 \\ \hline
CIFAR-10      & 90.81   & 90.65    & 91.40    & \textbf{91.46}    & 91.07    & 91.07     & 91.06     \\
CIFAR-100     & 64.25   & 64.82    & 65.77    & 66.44    & 66.08    & 66.43     & \textbf{66.82}     \\
STL-10        & 88.10   & 89.30    & 89.60    & 89.69    & 89.76       & \textbf{90.41}     & 90.11     \\
Tiny ImageNet & 44.82   & 45.93    & 47.04    & 47.31    & 47.74       & 48.09     & \textbf{48.34}    \\ \hline
\end{tabular}
}
\label{table: Queue_size}
\vspace{-1.0em}
\end{table}
\begin{table*}[htb]
\centering
\setlength{\belowcaptionskip}{6 pt}
\small
\caption{An analysis of the position of augmented views in MSVQ. The best results are indicated in bold, and the suboptimal results are underlined.}
\begin{tabular}{l|c|cc|cc|cccc}
\hline
Method & $X^2$ & $X^3~in~g_{t1}(f_{t1}(\cdot))$ & $X^3~in~g_{t2}(f_{t2}(\cdot))$ & $X^4$ & $X^5~in~g_{t2}(f_{t2}(\cdot))$ & CIFAR-10 & CIFAR-100 & STL-10 & Tiny ImageNet \\ \hline
MSVQ(Ours)   & \ding{51}   & \ding{51}         &          &  \ding{51}  &          &  \textbf{91.46}       &  {\ul66.44}         &  {\ul90.41}      & {\ul48.09}              \\ 
       & \ding{51}   &          &  \ding{51}        &\ding{51}    &          &  91.05        &  66.16         & 90.11       & 47.45              \\ 
       & \ding{51}   &  \ding{51}        &          & \ding{51}   & \ding{51}         & {\ul91.16}         & \textbf{67.14}          &  \textbf{90.49}      &  \textbf{48.55}             \\ 
MSV(Ours)      &  \ding{51}  & \ding{51}         &          &    &          &  90.92        & 65.02          & 89.35       & 46.68              \\ 
MQ(Ours)      &  \ding{51}  &          &          & \ding{51}   &          &  90.91        & 65.15          & 89.58       & 46.32              \\ 

ReSSL \cite{my_3zheng2021ressl}       & \ding{51}   &          &          &    &    & 90.20          & 63.79          & 88.25          & 46.60              \\ \hline
\end{tabular}
\label{table: position_of_the_view}
\end{table*}
\begin{figure*}
    \centering
    \begin{minipage}{0.32\textwidth}
        \vspace{+1.6em}
        \centering
        \includegraphics[width=\linewidth]{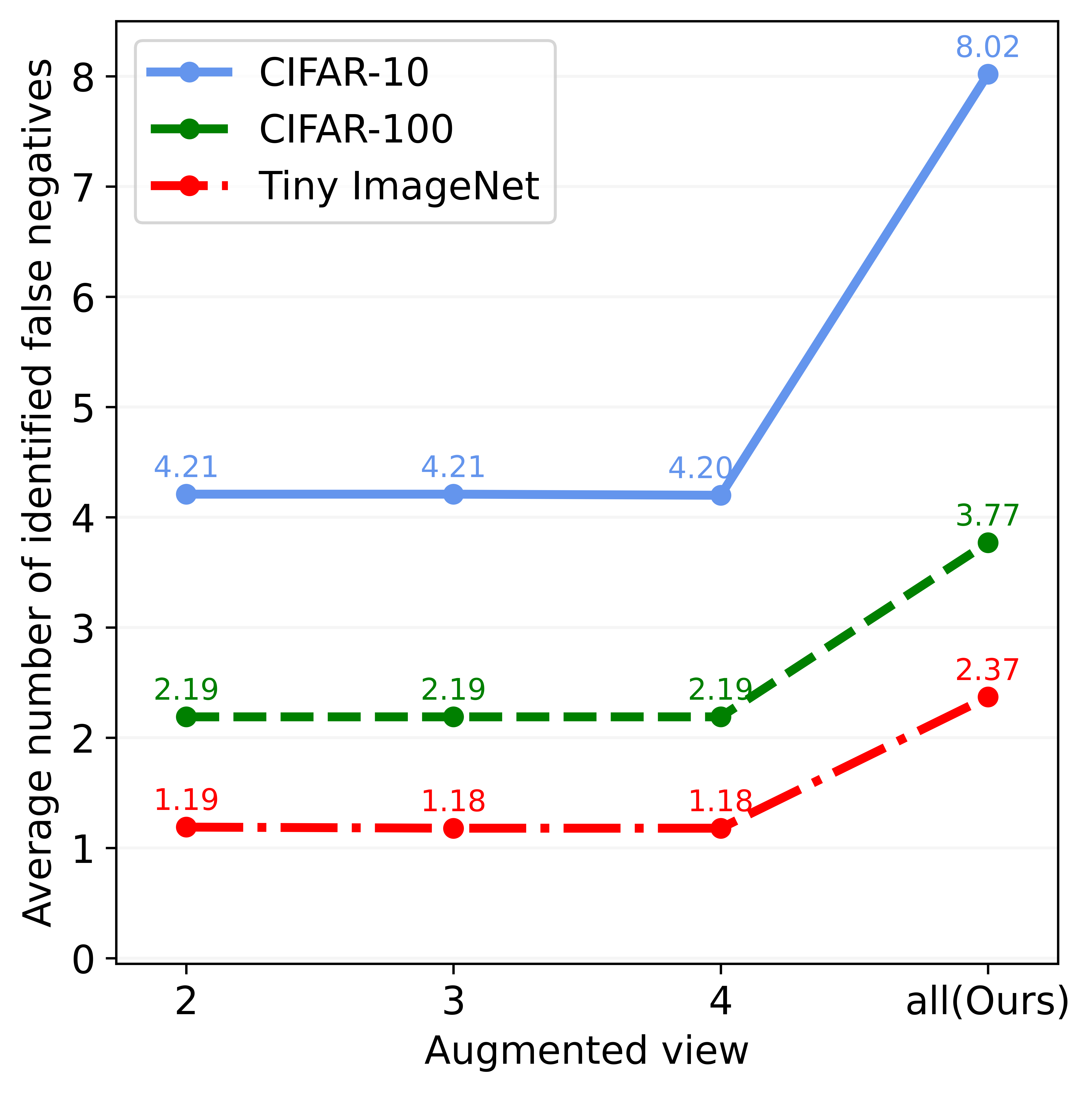} 
        \centering
        \caption{The average number of false negative samples identified by different augmented views (i.e., $X^2$, $X^3$, and $X^4$) in teacher networks. Here, 'all' represents the cumulative effect of all three soft labels when utilized simultaneously.}
        \label{figure: MSVQ_fns}
    \end{minipage}
    \hfill
    \begin{minipage}{0.64\textwidth}
        \centering
        \subfigure[MoCoV2]{
            \begin{minipage}[t]{0.5\linewidth}
            \centering
            \includegraphics[width=\textwidth]{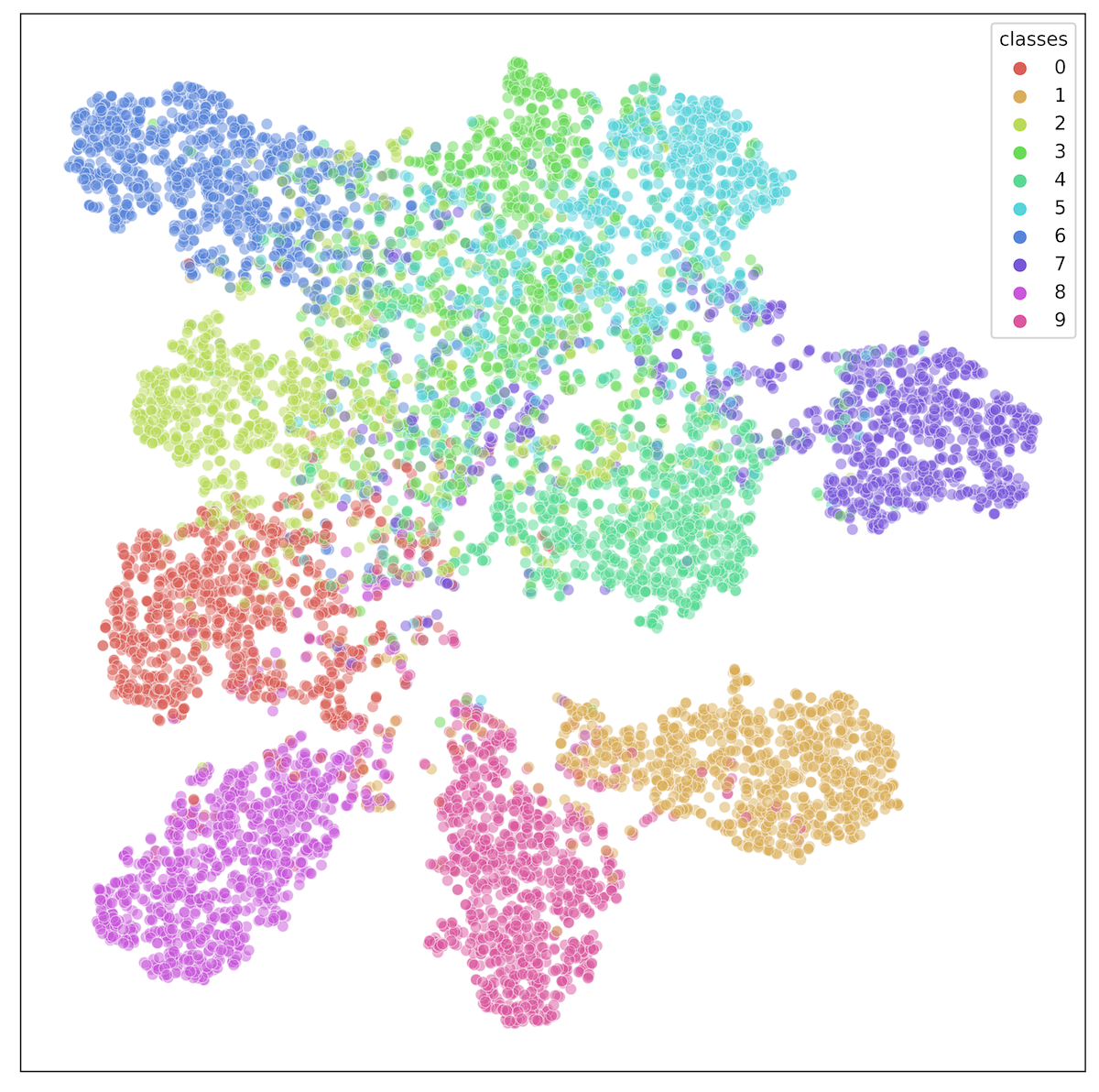}
		    \end{minipage}
	    }%
    	\subfigure[MSVQ]{
		    \begin{minipage}[t]{0.5\linewidth}
			\centering
			\includegraphics[width=\textwidth]{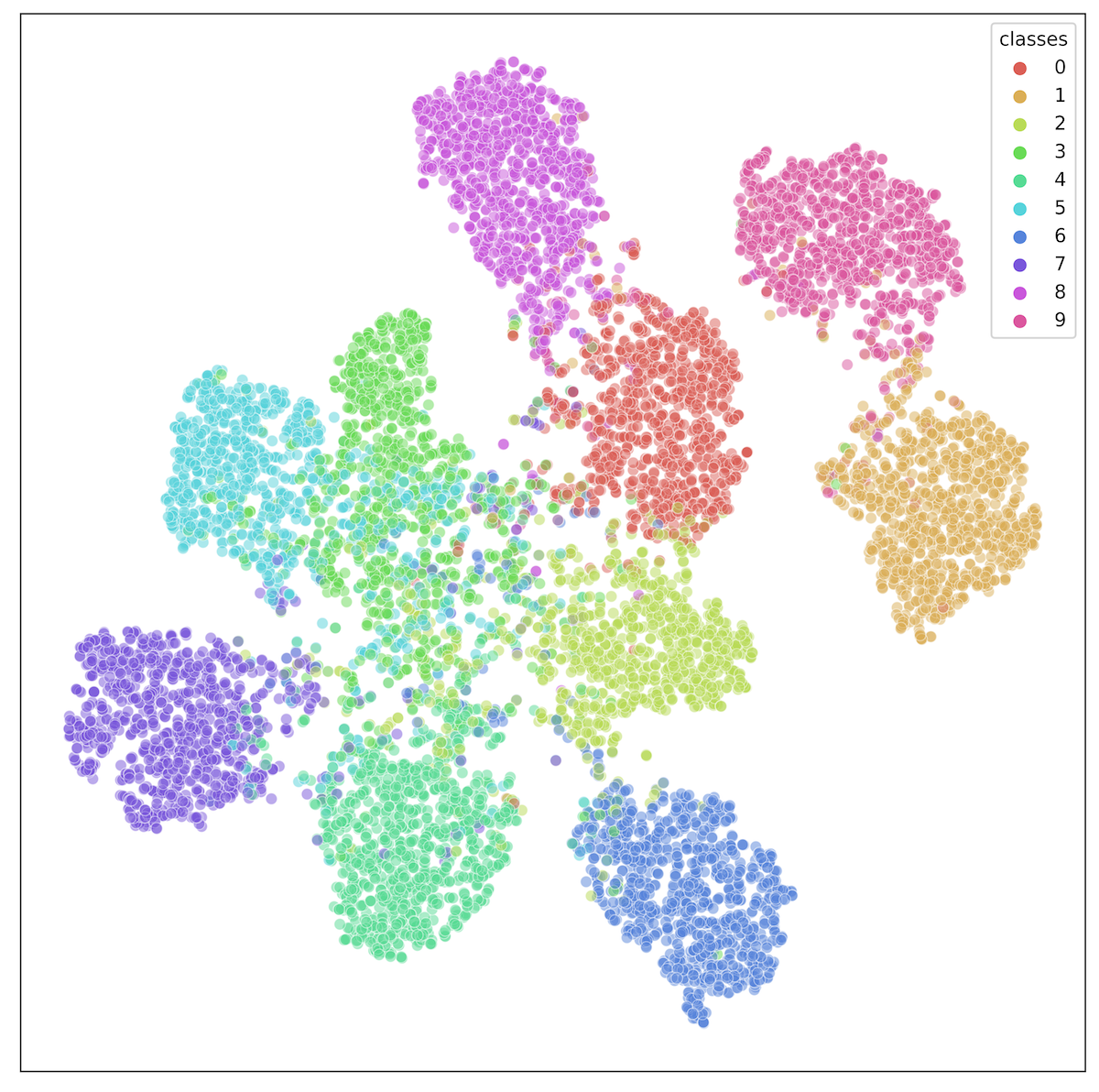}
		    \end{minipage}
	   }%
        \centering
	   \caption{t-SNE visualization of learned features on CIFAR-10, classes indicated by different colors. Best viewed in color.}
        \label{figure: t_sne}
    \end{minipage}
\end{figure*}
\subsection{Analysis and discussion}\label{46}
\subsubsection{The position of augmented views}\label{461}
In addition to the default configuration described in Sec.~\ref{323}, we conducted experiments exploring two variations of MSVQ: (i) relocating the augmented view $X^3$ from $g_{t1}(f_{t1}(\cdot))$ to $g_{t2}(f_{t2}(\cdot))$, and (ii) introducing a new view $X^5$ generated through weak augmentation in $g_{t2}(f_{t2}(\cdot))$. The experimental results, as presented in Table~\ref{table: position_of_the_view}, demonstrate that (ii) yielded slightly better performance improvements compared to (i). This suggests that the number of augmented views (i.e., distinct soft labels) may have a more significant impact on performance than the choice of the teacher network in which they are employed.

Furthermore, the default MSVQ settings yielded slight performance gains compared to variation (i) across all datasets. This was due to our experimental setup where we set the momentum update coefficient $m_1$ to be greater than $m_2$. Larger momentum update coefficients reduce disturbances caused by inconsistencies among different batches of negative samples in the queue \cite{24he2020momentum}. Given this simplicity and the optimal performance observed, we have chosen to retain the default settings of MSVQ.

\subsubsection{Analyzing model reliability and coverage in identifying false negative samples}\label{462}
We treat the distribution of relationships between ${\{X^i\}}_{i=2}^4$ and the negative samples in the queues as three distinct soft labels. These labels provide guidance for the student network in classifying the negative samples within the queues. Ideally, false negative samples should receive higher prediction values in the student network, and vice versa.

To ensure the accuracy of these three soft labels, we employ both weak data augmentation and lower temperature parameters in teacher networks. In this section, we investigate the reliability and scope of these three soft labels in the identification of false negative samples. These labels are expected to assign higher similarity values to the false negative samples. Fig.~\ref{figure: MSVQ_fns} illustrates the average number of false negative samples identified by the three soft labels. In detail, we begin by arranging each of these three distributions (i.e., $P^{2,1}$, $P^{3,1}$, and $P^{4,2}$) in descending order. Then, we calculate the average count of false negative samples that share the same labels as the positive sample within the \textit{top 5} samples of each distribution. In this context, 'all' refers to the total number of distinct false negative samples identified by aggregating their respective sets of false negative samples when utilizing all three soft labels simultaneously. This observation implies that the three soft labels in our model can effectively identify distinct sets of false negative samples within the queues, resulting in the recognition of nearly twice as many false negatives compared to when each soft label is applied individually.
\subsubsection{Visualization of features}\label{463}
As demonstrated by t-SNE visualization \cite{42van2008visualizing} in Fig.~\ref{figure: t_sne}, our method exhibits more distinct class boundaries and a more compact internal arrangement of classes compared to MoCoV2. This suggests that MSVQ offers a greater ability to alleviate the issue of false negative samples in the instance discrimination task.
\section{Conclusion}\label{5}
In this work, we bring in the framework of MSVQ. We improve the reliability and coverage of false negative sample identification by introducing two complementary and symmetrical methods to generate three distinct soft labels within the teacher networks. The first method entails utilizing multiple weakly augmented views of the positive sample, while the second method involves employing two momentum encoders to generate distinct semantic features for negative samples. Our extensive experimental results on four benchmarks demonstrate the remarkable performance of MSVQ. In future research, our goal is to explore even more effective strategies for leveraging semantic diversity within the realm of SSL.
\section*{Acknowledgement} \label{6}
This work was supported by the National Natural Science Foundation of China under Grant No. 61906098.
\bibliographystyle{elsarticle-num} 
\biboptions{sort&compress}
\bibliography{ref} 

\end{document}